# A Pareto Front-Based Multiobjective Path Planning Algorithm

Alexander Lavin

*Abstract*—Path planning is one of the most vital elements of mobile robotics. With *a priori* knowledge of the environment, global path planning provides a collision-free route through the workspace. The global path plan can be calculated with a variety of informed search algorithms, most notably the A* search method, guaranteed to deliver a complete and optimal solution that minimizes the path cost. Path planning optimization typically looks to minimize the distance traversed from start to goal, yet many mobile robot applications call for additional path planning objectives, presenting a multiobjective optimization (MOO) problem. Past studies have applied genetic algorithms to MOO path planning problems, but these may have the disadvantages of computational complexity and suboptimal solutions. Alternatively, the algorithm in this paper approaches MOO path planning with the use of Pareto fronts, or finding non-dominated solutions. The algorithm presented incorporates Pareto optimality into every step of A* search, thus it is named A*-PO. Results of simulations show A*-PO outperformed several variations of the standard A* algorithm for MOO path planning. A planetary exploration rover case study was added to demonstrate the viability of A*-PO in a real-world application.

## I. Introduction

A crucial task for mobile robots is to navigate intelligently through their environment. It can be argued that path planning is one of the most important issues in the navigation process [1], and subsequently much research in field robotics is concerned with path planning [2], [3]. To complete the navigation task, methods read the map of the environment and search algorithms attempt to find free paths for the robot to traverse. Path planning methods find a path connecting the defined start and goal positions, while environmental parameters play the role as algorithm inputs, and the output is an optimized path from the start to goal [4]. An important issue in mobile robot navigation is optimizing path efficiency according to some parameters such as cost, distance, energy, and time. Of these criteria, time and distance are typically the most important for researchers [5], and methods typically optimize the path efficiency for only one criterion [6]. Yet many mobile robot operations call for a path plan that is efficient over several parameters. Path optimization over several parameters – e.g. distance and energy – is a multiobjective optimization (MOO) problem. The best path is not necessarily the shortest path, nor the path calling for the least amount of energy expenditure.

Combining the optimization criteria into a single objective function is a common approach, often with tools such as thresholds and penalty functions, and weights for linear combinations of attribute values. But these methods are problematic as the final solution is typically very sensitive to small adjustments in the penalty function coefficients and weighting factors [6]. Evolutionary algorithms, particularly genetic algorithms, have been used widely for MOO problems, including success in path planning [7], [8]. Some state-of-the-art algorithms for multi-objective evolutionary computation include NSGA-II and SPEA2 [9], [10]. The merging of path segments can result in offspring solutions with high scores across several fitness criteria. The *non-dominated* paths are favored in the population, and this increases generation over generation [11]. Non-dominated solutions are those in which there exists no other solutions superior in all attributes. In attribute space, the set of non-dominated solutions lie on a surface known as the *Pareto front*. Fig. 1 illustrates the two-dimensional case, where there is a tradeoff between minimizing both $f_1$ and $f_2$. The goal of a Pareto evolutionary algorithm is to find a set of solutions along the Pareto front, optimal for a combination of criteria [12].

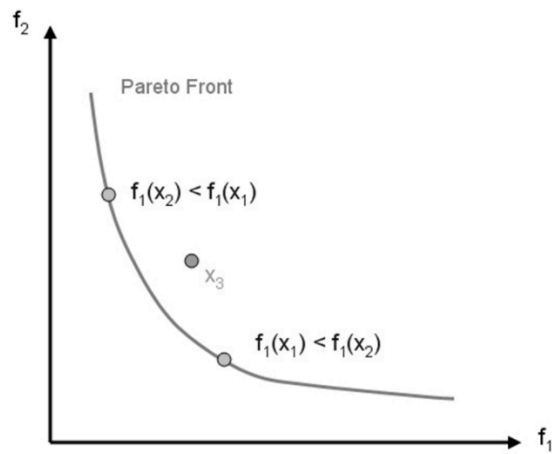

Fig. 1. Two-dimensional Pareto space, where points $x_1$ and $x_2$ lie on the Pareto front [13].

This study looks to use multiobjective optimization for mobile robot path planning, but with a Pareto front cost function. Other studies have applied Pareto optimality to evolutionary planning for synchronous optimization of several objectives [8], and domination metrics are used in some evolutionary algorithms for path planning, including NSGA-II and SPEA2 [11], [12]. Yet these algorithms compare complete paths for domination. In order to sort a population according to the level of non-domination, each path must be compared with every other path in the population to find if it is dominated, where the computational complexity scales exponentially with the search space [9]. The algorithm presented in this study, however, checks for non-domination at each search step, resulting in a single, optimal path. The path planning algorithm is novel because each step is Pareto optimal.

The next section further discusses Pareto optimality and the application to mobile robot path planning. Section III discusses the technical approach used in this study, and Section IV presents the results. Included in Section IV is a Mars rover case study as an example application of the new A*-PO algorithm. Other applications for mobile robots with global path planning include agricultural harvesting and information gathering (i.e. drones), disaster relief, DARPA challenges, factory and residential robot workers, and exploration rovers. Section V concludes the paper with discussion and future work.

## II. MATERIALS AND METHODS

### A. Mobile Robot Path Planning

The aim of mobile robot path planning is to provide an efficient path from start to goal that avoids objects and obstacles. An efficient path is one that minimizes path costs, where the cost is typically the travel distance or time.

Path planning methods can be categorized as either *static* or *dynamic*, according to the environmental conditions. The positions of all obstacles and objects in the static environment are fixed and known. The dynamic environment, on the other hand, may have obstacles and objects that vary positions with time. Similarly, an unknown environment calls for dynamic path planning because more is learned as the mobile robot progresses through the environment. The algorithms for path planning are also in two categories: *local* and *global*. Local algorithms function as the robot moves through the environment, revising the path based on environmental changes. Global algorithms use *a priori* knowledge of the environment to plan the path, and are thus applicable to planning in static environments. Each method has its own pros and cons depending on the environment and application type [8].

The control architecture in mobile robotics is typically a combination of local and global planners, organized as shown in Fig 2. The reactive layer handles local information, with real-time constraints. The deliberative, or global, layer considers the entire world, likely requiring computation time proportional to the problem size [15]. The algorithm presented in this paper is a global path planner.

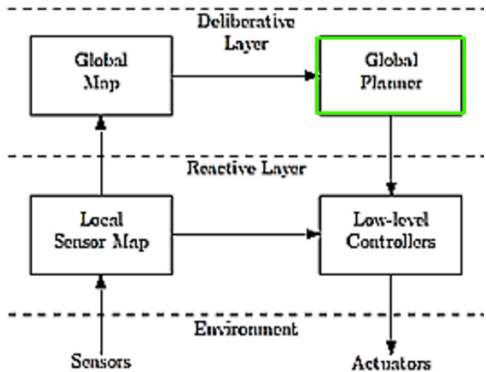

Fig. 2. High-level block diagram of the standard hybrid control system architecture for mobile robots [14]. The focus here is global path planning.

There are two main components of global path planning. First is the robot representation of the world in the *configuration space*: data structures that show the position and orientations of objects and robots in the workspace area, including both the free and obstructed regions. The configuration spaces of path planning algorithms are usually represented by an occupancy grid, a vertex graph, a Voronoi diagram, generalized cones, or a quad-tree [1].

The methods discussed in this study use an *occupancy grid*, where the environment is represented by a two-dimensional layout of square cells. The values of these cells are binary states, where 0s and 1s represent free and occupied spaces, respectively. The robot occupies a cell, with or without orientation. For a given cell currently occupied by the robot, there are eight feasible cells in the path that can be successors. This is shown in Fig. 3, where the robot in the green position is capable of moving into a neighboring yellow position, but not the occupied gray cells. Feasible solution paths never collide the robot with an obstacle.

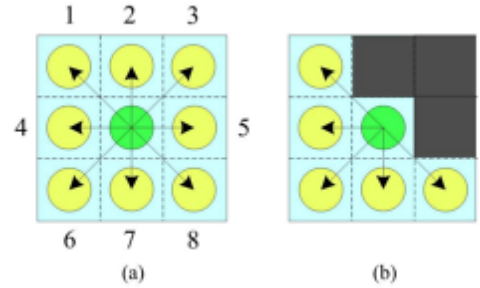

Fig. 3. (a) The robot (green cell) has at most eight possible path steps. (b) The set of feasible successor cells is narrowed because of the three occupied cells (gray) [9].

The second main component of global path planning is implementing an algorithm to find an optimal path from start to goal states. That is, for two arbitrary points in the area – the start and the goal – the algorithm finds a drivable path between them that minimizes distance, energy, or some other criteria. The algorithm employed for the problem must coordinate with the configuration space representation [1]. Potential solution paths connect the start cell to the goal cell via free cells. Searching for the most efficient path is an optimization problem, where the optimum path is defined as that which minimizes the path cost, or the *objective function*.

A candidate path can be denoted by

$$P = \{p_1, p_2, \ldots, p_n\} \quad (1)$$

where $p_i$ is the $i$th waypoint of the path $P$. The MOO problem is then framed as determining a path

$$P^* \in P \quad (2)$$

that satisfies

$$F(P^*) = min\{F_1(P), F_2(P), \ldots, F_t(P)\} \quad (3)$$

where $F_i$ denotes the $i$th cost function of the path planning problem. The study here considers three cost functions, or $t = 3$. They are defined in (4) and (5) below, and (6) later in the Mars rover case study.

Equation (4) gives the total length of the path:

$$F_1(P) = \sum_{i=1}^{n-1} |p_i, p_{i+1}| \quad (4)$$

where $|p_i, p_{i+1}|$ is the Euclidean distance between subsequent cells in the path. Minimizing $F_1$ finds the path of shortest length from start to goal.

Equation (5) gives the average elevation of the path:

$$F_2(P) = \sum_{i=1}^{n} e_i / n \quad (5)$$

where $e_i$ is the elevation at waypoint $i \ldots n$. With the fixed start and goal states at constant elevation, the minimization of $F_2$ gives the path that climbs up the least amount of incline (or alternatively moves the robot down the most decline).

Search algorithms are employed for finding the minimal cost paths through the configuration space. *Uninformed search* methods are used when no information about the states are known beyond the problem definition [14]. The global path planning problem discussed here has *a priori* knowledge – a map of the exploration area. Thus uninformed search methods, like Dijkstra's breadth-first algorithm, can be ignored in favor of *informed search* methods. The general approach of these methods is *best-first*, which traverses a graph or grid using a priority queue to find the shortest, collision-free path [4]. The decision of the next node expanded, the *successor*, is based on an *evaluation function*, $f(n)$: estimated cost of the cheapest solution through node $n$. The choice of $f(n)$ determines the search strategy. A bonus of informed search is including a *heuristic function* $h(n)$: the estimated cost of the cheapest path from a node $n$ to the goal state. Greedy best-first search is built solely on this heuristic, where $f(n) = h(n)$, expanding the node closest to the goal at each search step. The incorporation of the heuristic into the path cost makes the search algorithm more efficient. The search algorithm also gains efficiency by using a *priority queue*, or *open list*, to update the costs of nodes. From the open list, the algorithm chooses successor nodes to expand.

The A* algorithm is perhaps the most popular best-first search method, adding to the heuristic the cost to reach the node, $g(n)$. That is, $f(n) = h(n) + g(n)$. The search algorithm, looking for the cheapest path, tries (expands) the node with the lowest $f(n)$ [15], [16]. To determine the optimal sequence of waypoints (i.e. path), the A* algorithm is a favorite for route search problems [17], [18]. For graph search, as opposed to tree search, a *consistency condition* is required to guarantee optimality. A heuristic is consistent if, for every node $n$ and every successor $n'$ of $n$ generated by any action $a$, the estimated cost of reaching the goal from $n$ is no greater than the step cost of getting to $n'$ plus the estimated cost of reaching the goal from $n'$:

$$h(n) \leq c(n, a, n') + h(n') \quad (6)$$

Norvig and Russel [15] explain how the A* heuristic satisfies the consistency condition, and also that A* is *optimally efficient*: no other optimal algorithm is guaranteed to expand fewer nodes than A*. As long as a better-informed heuristic is not used, A* will find the least-cost path solution at least as fast as any other method.

For real-time planning, where computational speed is a priority, previous studies [19], [20] have modified A* for fast planning. The D* algorithm is a dynamic version of A*, built to be capable of fast rerouting when the robot encounters new obstacles in the environment [4]. The speed of these searching algorithms is increased dramatically, but at the cost of sub-optimal solution paths [14].

*B. Pareto Optimality*

The MOO problem presents multiple cost criteria, where a solution stronger for one criterion may be weaker for another. There are two general approaches to optimizing for multiple objectives: (i) combine the individual objectives into one composite function, and (ii) determine a *Pareto optimal* solution set. The first can be accomplished with weighted sums or utility functions, but selection of parameters is difficult because small perturbations in the weights can lead to very different solutions. The second option finds the Pareto optimal set of the population, which is a set of solutions that are non-dominated with respect to each other. That is, moving between Pareto solutions, there is always sacrifice in one objective to achieve gain in another objective [21]. It is advantageous to incorporate Pareto fronts in evolutionary algorithm fitness functions when tackling MOO problems. Simply summing over the fitness criteria presents difficulties. Yet in search methods it is common the cost function sums over the cost criteria at each step; the A* algorithm sums $h(n)$ and $g(n)$.

| **Algorithm 1** A* Search |
|---|
| 1 **Initialize** open and closed lists |
| 2 Put the starting node in the open list |
| 3 Define f, the cost function |
| 4 **While** the open list is not empty |
| 5     q ← node on open list with smallest f |
| 6     Remove q from open list |
| 7     Generate q's 8 successors, set their parents to q |
| 8     **For** each successor |
| 9        **If** successor is a goal, then stop search |
| 10       successor.g ← q.g + distance between successor and q |
| 11       successor.h ← distance from successor to goal |
| 12       successor.f ← successor.g + successor.h |
| 13       **If** a node with same position as successor is in the open list & has a lower f than successor, then skip this successor |
| 14       **If** a node with same position as successor is in the closed list & has a lower f than successor, then skip this successor |
| 15       **Else**, add the node to the open list |
| 16     **End For** |
| 17     Push q to the closed list |
| 18 **End While** |

For minimization of objective function $f$, a point $n^*$ is said to be a Pareto optimal point if there is no $n$ such that $f_i(n) \leq f_i(n^*)$ for all $i = 1 \dots t$, where there are $t$ optimization objectives.

Point $n^* \in C$ is a non-inferior solution if for some neighborhood of $n^*$ there does not exist a $\Delta n$ such that $(n^* + \Delta n) \in C$,

$$f_i(n^* + \Delta n) \leq f_i(n^*), i = 1, \dots, m, \text{ and}$$

$$f_j(n^* + \Delta n) \leq f_j(n^*) \text{ for at least one } j.$$

Multiobjective optimization is, therefore, concerned with the generation and selection of non-inferior solution points – those on the Pareto front. Pareto optimality is a crucial concept for finding solutions to MOO problems because identifying a single solution that simultaneously optimizes across several objectives is often an impossible task [22].

It is worth noting that summing over the costs to calculate a composite $f$ presents another possible issue in search algorithms: depending on the current development of the path, some cost criteria may be favored over others, and this changes as the path development continues. For instance, the A* heuristic – the estimated cost of the cheapest path from the current cell to the goal cell – will contribute more to the cost function close to the start than it will close to the goal. That is, near the start state $h(n)$ will have greater influence on $f$ than will $g(n)$; the inverse is true near the goal state. Thus, as the path develops from start to goal, the heuristic value will contribute less and less. *Using a Pareto front solves this issue because each cost criterion is valued as its own dimension in the Pareto space, not summed together*.

### III. TECHNICAL APPROACH

#### A. Costmap

To calculate cost functions at each step the search algorithm uses a *costmap*. This representation of the configuration space is built off of the aforementioned occupancy grid, but now a cost value is assigned to each cell. Traversing a free space adds a unit cost to the path total, and the obstacles are represented by infinite cost; thus, they are not traversable. If traversing straight across a cell carries a unit distance cost, the cost for traversing a cell at a diagonal (a 45° angle) carries a cost of $\sqrt{2}$.

Yet this costmap only reflects the distance of taking a given path through the configuration space. For a MOO problem, the path cost needs to consider the other cost criteria, for which I use additional *layers*. Each additional cost layer adds a dimension to the Pareto space, from which the Pareto front is calculated. The first costmap layer is the distance cost, $g(n)$. The second layer is the heuristic, $h(n)$. These two suffice for traditional A* search, but I'm also interested in optimizing the robot's path for elevation – i.e. minimize (5). A third layer, $e(n)$, is then added to the costmap. With three layers, the Pareto space is three-dimensional. That is, *points on the Pareto front are optimal across the three dimensions, one for each cost – distance, the heuristic, and elevation*.

**Algorithm 2** A*-PO Search (replaces line 8+ of Alg. 1)
8  **For** each successor
9     **If** successor is a goal, then stop search
10    successor.g ← q.g + distance between successor and q
11    successor.h ← distance from successor to goal
12    successor.e ← elevation of succesor
13    scoreMatrix(successor) ← [successor.g, … successor.h, successor.e]
14 **End For**
15 q ← Calculate Pareto front of scoreMatrix
16 **If** multiple points on Pareto front
17    Normalize scoreMatrix
18    q ← run std. A* cost function on Pareto front nodes
19 Push q to the closed list
20 **End While**

#### B. A*-PO Search Algorithm

The algorithm presented in this study, A*-PO, is essentially the standard A* search algorithm but for a key modification: rather than computing the cost function $f$ by summing cost criteria, A*-PO calculates the Pareto front of the cost criteria. Lines 8-16 in the previous A* pseudocode are replaced by the pseudocode shown above.

Calculating the Pareto front of the open list will at times yield multiple Pareto points. That is, q (in the pseudocode) may contain multiple Pareto points. For this scenario, where multiple nodes makeup the Pareto front, one is chosen from the set of Pareto points via the normalized A* cost calculation. For instance, consider a given step of the path search with an open list (i.e. priority queue) consisting of 11 nodes, and perhaps three fall on the Pareto front. The algorithm will first normalize the three nodes for each cost criteria such that the range for each criterion is [0:1] for the set of nodes on the Pareto front; note the normalization is across each dimension of the Pareto space. Then the A* cost metric is used to decide between these Pareto front nodes. Thus, the A*-PO search algorithm still maintains the quality that every step is Pareto optimal; i.e. *the successor node is always a Pareto optimal point*.

### IV. RESULTS

The MOO path planning algorithms are tested in simulated mobile robot environments. The computer simulation environment includes a Lenovo notebook computer with Intel Core i5 vPro CPU and 4 GB memory, running on Windows 8.1. The code is written in MATLAB R2013a, and will be published in a future release of Corke's "Robotics, Vision, Control" [23].

#### A. Algorithm Comparison

The A* search algorithm is guaranteed complete and optimal, but not necessarily for MOO path problems. The advantages of A*-PO are significant for MOO path problems. I evaluated A*-PO by comparing it with A* for a set of 80 simulated environments.

The workspaces were setup as 20x18 cell grids of randomly assigned free spaces and obstacles, the obstacles accounting for 20% of the workspace. The start and goal locations were fixed at the upper left (0,0) and lower right (20,18), respectively. The elevations for the goal and start states were at 0 in each in configuration, where the terrain ranged [0:1]. Eight unique terrains were used in the simulations.

The optimization objectives, as presented above in (3) and (4), were to minimize the total path distance and elevation. Fig. 4 shows an example of the resulting paths for each search method in pink, where the gray squares represent the path steps; the red and green marked squares represent the start and goal states, respectively. The left side diagrams of Fig. 4 are the final solution paths over a grid of obstacles (black) and free spaces (white), representing the occupancy grid layer of the costmap. The right side diagrams show the same paths over a contour map, representing the elevation layer of the costmap.

For one of the 80 simulation runs, Fig. 4a shows the final solution path of the standard A* algorithm for the distance travelled, the heuristic, and the elevation cost criteria. At each search step the costs for each criteria were normalized [0:1] over the nodes in the open list. It was necessary to normalize the path costs at each search step because the elevation values are small relative to the distance values; without normalization the elevation metric would be insignificant. This normalization is unnecessary for the A*-PO algorithm because each cost value is relative to the cost metric's dimension in Pareto space. Fig. 4b shows the solution path for the A*-PO algorithm of the same simulation environment as A* in Fig. 4a.

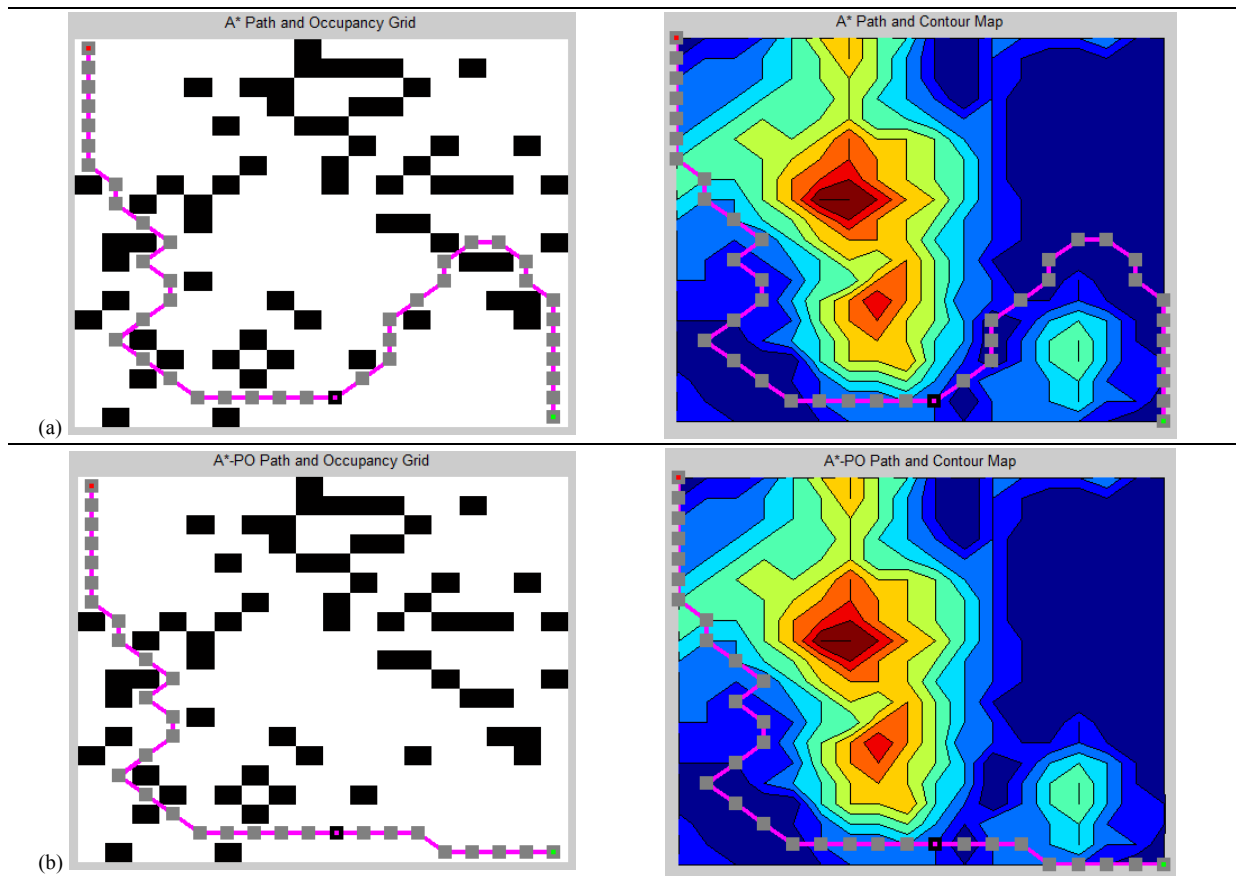

Fig. 4. Solution paths for (a) A* and (b) A*-PO from one of the 40 simulation runs., where the path cost at each step includes distance travelled, the heuristic, and elevation. The divergence in the two paths is plotted with a black square.

| Algorithm | Environment | $F_1$: Path Length (steps) | $F_2$: Average Elevation [0:1] | $F_3$: Solar Incidence [0:1] | Search Time (seconds) |
|---|---|---|---|---|---|
| A* | Simulation | 31.1 | 0.72 | n/a | 0.19 |
| A*-PO | Simulation | 28.3 | 0.65 | n/a | 0.28 |
| A* | Mars | 161.4 | 0.45 | 0.53 | 1.21 |
| A*-PO | Mars | 151.3 | 0.39 | 0.69 | 1.42 |

Table 1 – Results of simulations and case study

For the sample workspace and terrain in this example, it is clear to see the benefits of calculating the Pareto front at each search step. The data over the set of 80 simulations echo these results, as shown in Table 1. The A*-PO algorithm outperforms the other A* variations for the optimization objective functions $F_1(P)$ and $F_2(P)$, the path length (steps) and average elevation (normalized), respectively. The search time results show paths with Pareto optimal steps can be obtained efficiently with the A*-PO algorithm, with only a slight increase in computation time over the standard A* search algorithm. All algorithms gave complete solution paths.

The average elevation of each solution path is used as a metric to compare the robot's net incline from start to goal. A path of a given average elevation implies the robot traversed up less slope (or down more slope) as compared to a path of higher average elevation. Lower values are preferred for the mean elevation, as well as the other path cost metrics.

### B. Case Study

A case study is presented to demonstrate the application of the A*-PO algorithm in a potential use case. An example Mars terrain was sourced from HiRISE, the High Resolution Imaging Science Experiment conducted by the University of Arizona, NASA, JPL, and USGS [24]. Fig. 5 shows a digital terrain model of a Mars landscape, from which a section (red square) was extracted for use in the case study.

The extracted section was converted to a terrain map with elevation values [0:1], as shown in Fig. 6. The overlaid occupancy grid was generated randomly, with obstacles accounting for 30% of the workspace. The dimensions are 100x100, where each cell represents a 1m² area.

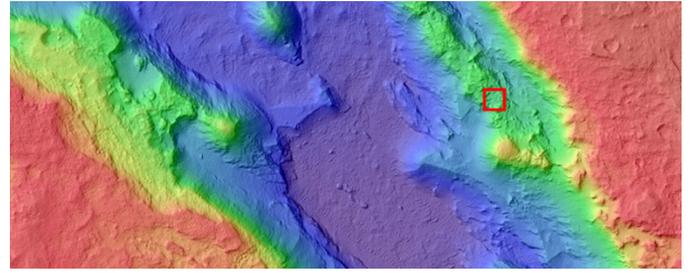

Fig. 5. Terrain map showing exposures of layered bedrock northwest of the Hellas Region of Mars. The selection within the red square was used for simulation.

In addition to the path planning objectives used above, the case study included an additional aim of maximizing the solar incidence on the rear of the rover. That is, the MOO problem included an additional optimization objective to minimize the total angular deviation of sunlight from the solar panel. This was computed by minimizing the dot product of the rover vector $\vec{r}$ and the solar ray vector $\vec{s}$:

$$F_3(P) = \vec{r} \cdot \vec{s} = |r||s|cos\theta \qquad (7)$$

The solar incidence cost criteria was incorporated as an additional layer to the costmap. However, this layer was dependent on the robot's orientation in the configuration space, and was thus dynamic. That is, the costmap changed at each step in the path, depending on the two-dimensional rover vector. For this case study, the solar angle was held constant and two-dimensional. Thus, there were eight variations of the solar costmap, or one for each possible angle between the solar and rover vectors.

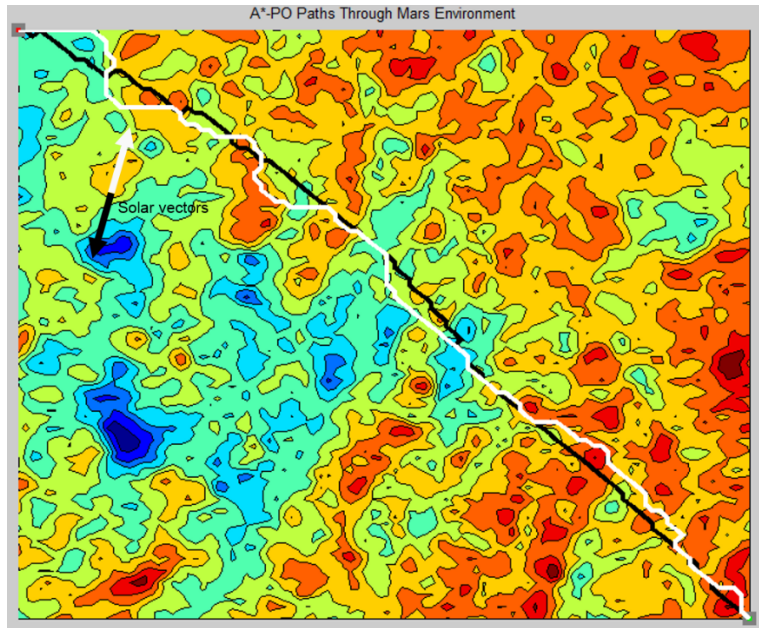

Fig. 6. The A*-PO solution paths through the Mars workspace (total elevation variation is 34.4m). The two paths both minimized the criteria for distance travelled, the heuristic, elevation, and solar angle deflection. The white path reflects solar incidence at angle 70°, while the black path is for 250°.

The result paths shown in Fig. 6 are Pareto optimal at each step across three independent cost functions, $F_{1-3}$ of (4), (5), (7). The three functions cover the four cost criteria because both the distance travelled and the heuristic contribute to $F_1$. The case study shows the A*-PO algorithm provides the least-cost global path according to several independent preferences for a mobile robot in practice.

Further studies may aim to more accurately include the solar incidence as a cost metric. This can be done by varying the angle of sunlight with time, as the rover progresses along its path. Or calculating the solar incidence in three-dimensional space. Additionally, one may account for more elaborate thermal constraints, such as heating of sensitive components by direct sunlight.

## V. Conclusion

In this study, global path planning for mobile robots is investigated. The optimal path is generated according to several cost criteria, solving the multiobjective optimization problem with the presented A*-PO algorithm. As demonstrated in the simulations, A*-PO is capable of providing paths where each step is Pareto optimal, and computes these solutions efficiently. In comparison to the traditional A* algorithm in this study, it can be concluded the use of Pareto fronts in A*-PO offers a better MOO search algorithm.

In future work, Pareto optimality may be incorporated into other algorithms of the mobile robot control system architecture (Fig. 1). The mobile robot community has put an increased emphasis on suboptimal path planning methods which meet the time-critical constraints over slow, optimal algorithms [14]. Local and dynamic path planners, such as D*, may improve with Pareto cost functions.

## Acknowledgment

The author would like to thank Professors Matthew Eicholtz and David Wettergreen of Carnegie Mellon University for their continued support and mentorship.